\documentclass[runningheads]{llncs}

\usepackage[table,x11names]{xcolor}
\usepackage{graphicx}
\usepackage{tikz}
\usepackage{comment}
\usepackage{amsmath,amssymb} 
\usepackage{color}

\begin{document}
\pagestyle{headings}
\mainmatter
\def\ECCVSubNumber{11}  

\title{A Hybrid Approach for 6DoF Pose Estimation} 


\titlerunning{Hybrid 6DoF Pose Estimation}
%
\author{Rebecca K\"onig\orcidID{0000-0002-4169-6759} \and
Bertram Drost\orcidID{0000-0002-4109-6999}
}
\authorrunning{K\"onig and Drost}
%
\institute{MVTec Software GmbH, Munich, Germany\\
\email{\{koenig,drost\}@mvtec.com}}
\maketitle

\begin{abstract}
We propose a method for 6DoF pose estimation of rigid objects that uses a state-of-the-art deep learning based instance detector to segment object instances in an RGB image, followed by a point-pair based voting method to recover the object's pose. We additionally use an automatic method selection that chooses the instance detector and the training set as that with the highest performance on the validation set. This hybrid approach leverages the best of learning and classic approaches, using CNNs to filter highly unstructured data and cut through the clutter, and a local geometric approach with proven convergence for robust pose estimation.
The method is evaluated on the BOP core datasets where it significantly exceeds the baseline method and is the best fast method in the BOP 2020 Challenge.

\end{abstract}

\section{Introduction}

Many of the recently published approaches for 6DoF object detection, especially on the BOP benchmark~\cite{hodan2018bop}, follow a two-stage pipeline. The first stage is a state of the art deep learning object detector that outputs the potential locations of object instances, often as bounding boxes or instance masks. The second stage iterates over those instances and, for each, estimates the instance's pose.

The main technical differences in the approaches are in the second, pose estimation stage. The deep-learning based approaches can roughly be categorized by the type of data representation they use (mostly image-based convolutions~\cite{park2019pix2pose,Sundermeyer_2020_CVPR,sundermeyer2020augmented,li2019cdpn} vs. Graph-based networks~\cite{gao2018occlusion}), and the kind of pose estimation they employ (direct regression of pose parameters~\cite{gao2018occlusion}, regression of 3D model coordinates of the scene points usually followed by PnP~\cite{park2019pix2pose,li2019cdpn,DBLP:journals/corr/abs-1902-11020,hodan2020epos}, or employing a codebook for estimating the rotation of an object~\cite{Sundermeyer_2020_CVPR,sundermeyer2020augmented}).

A key observation from the BOP 2019 challenge was that while point-pair voting based methods~\cite{drost2010model,vidal2018method}, which are not learning based, overall had the highest recognition rates, they were also among the slowest methods. One reason is the large search space, as the voting is performed on the complete scene without pre-segmentation or pre-detection of instances. Well-trained deep-learning based instance segmentation methods, on the other hand, estimate the locations of the objects in a scene rather fast.

We therefore combine the two approaches and perform the point-pair voting of~\cite{drost2010model} only on locations returned by an instance segmentation network. This combines the advantages of both methods: the deep network's ability to quickly filter through complex real-world data and to narrow down the search space, and the provable robustness of the point-pair voting for recovering the pose.

\section{Method}

\begin{figure}[t]
\centering
\includegraphics[width=0.9\linewidth]{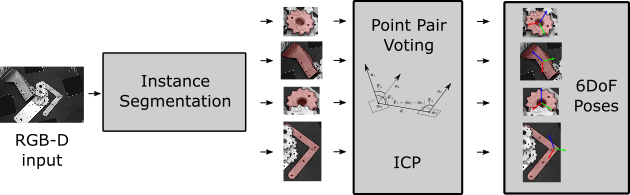}
\caption{Overview of the method, which follows a classical two-stage approach. First, a state-of-the-art network is used to find object instances (masks and object IDs) in the RGB image. Second, the point-pair voting is used to recover an object pose for each instance, using the depth image.}
\label{fig:method}
\end{figure}

We use a detection pipeline that uses a Deep-Learning based instance segmentation method as first stage, which returns regions and class IDs, followed by a point-pair voting in the regions of the detected instances as second stage. To account for the large domain differences in the BOP datasets, we automatically select the best instance segmentation method and training dataset based on the performance on the validation set.

\paragraph{Instance Segmentation}
The datasets in the BOP benchmark cover a variety of different object types and object placements. The objects have different geometric features and textures, while the placements range from isolated single objects to cluttered, unordered heaps of objects of the same instance. We found that a single object detector does not always cover all those cases properly. Instead, we train for each dataset a Mask-RCNN~\cite{he2017mask} and a RetinaMask~\cite{fu2019retinamask} network and automatically select the detector with the highest mean average precision (mAP) with an Intersection over Union (IoU) threshold at 0.5 \cite{everingham2010pascal} on the validation set. We use Mask-RCNN for the datasets YCB-V, T-LESS and ITODD, and RetinaMask for LM-O, HB, TUD-L and IC-BIN. We assume that Mask-RCNN performs better on these datasets since they have many classes of which some are very similar to each other. The two-stage approach of Mask-RCNN is probably better suited for these kind of datasets whereas RetinaMask directly classifies the anchor-boxes.

\begin{figure}
    \centering
    \includegraphics[width=0.24\linewidth]{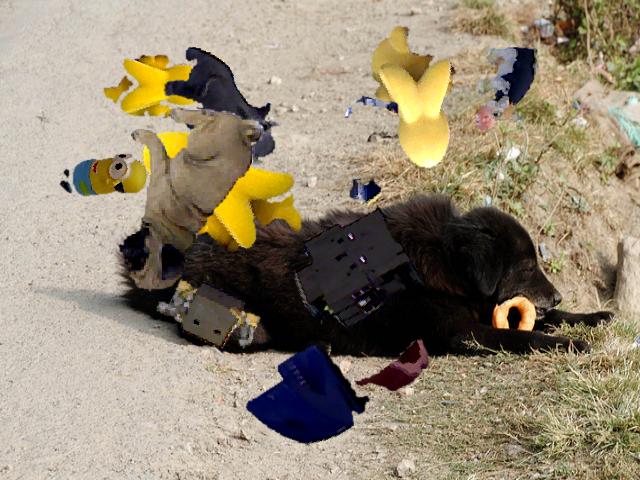}
    \includegraphics[width=0.24\linewidth]{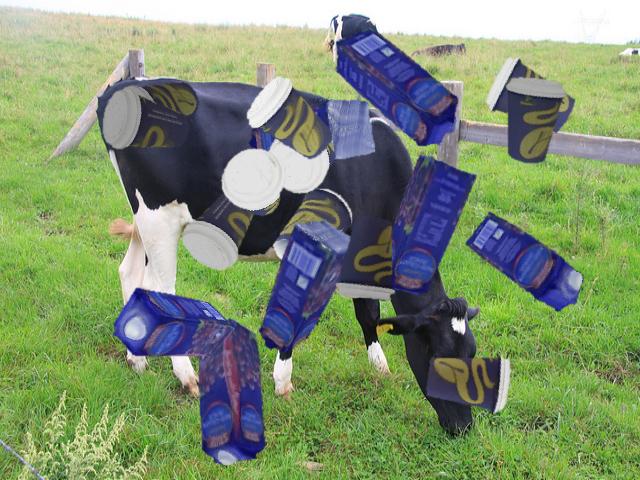}
    \includegraphics[width=0.24\linewidth]{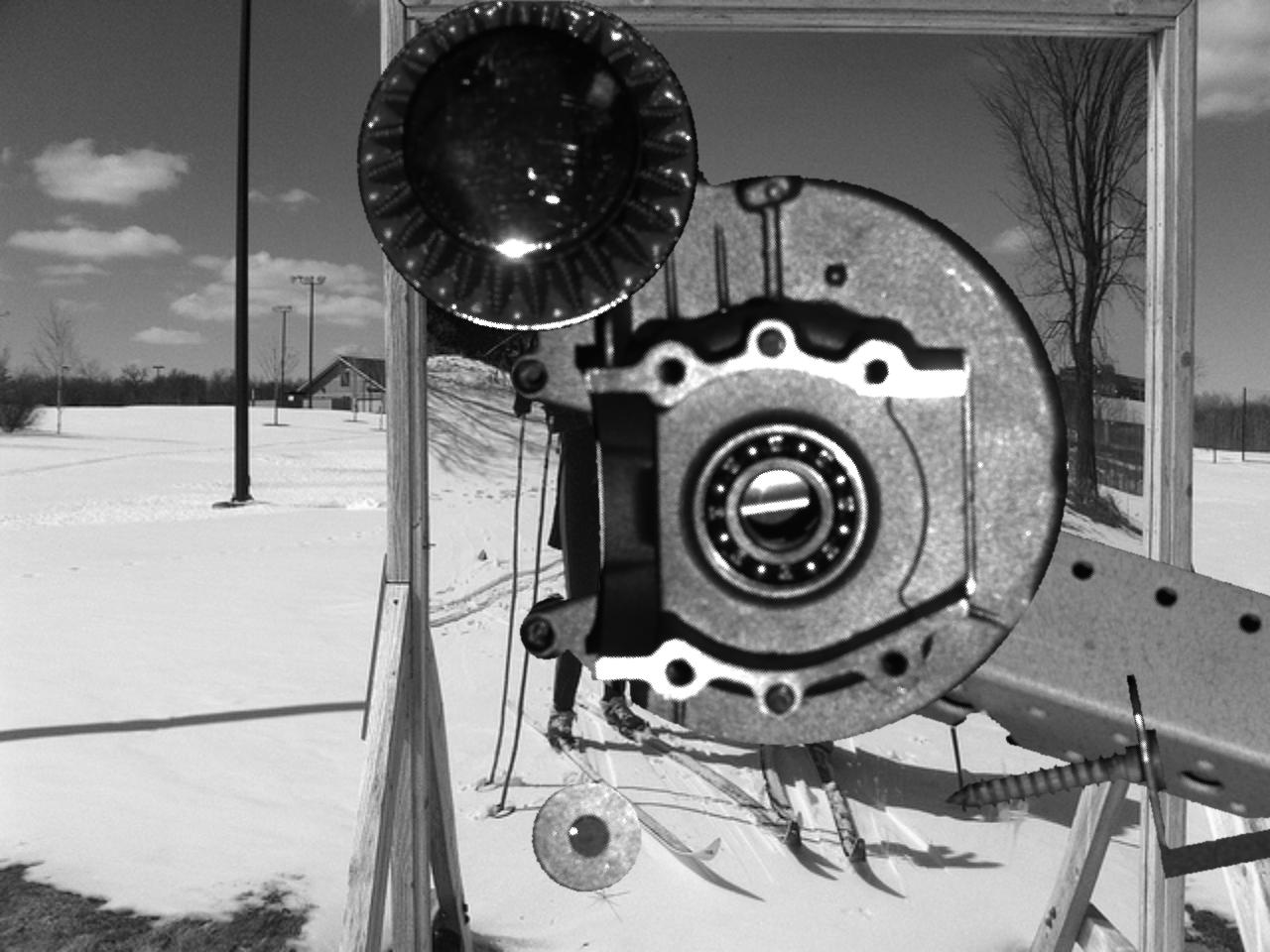}
    \includegraphics[width=0.24\linewidth]{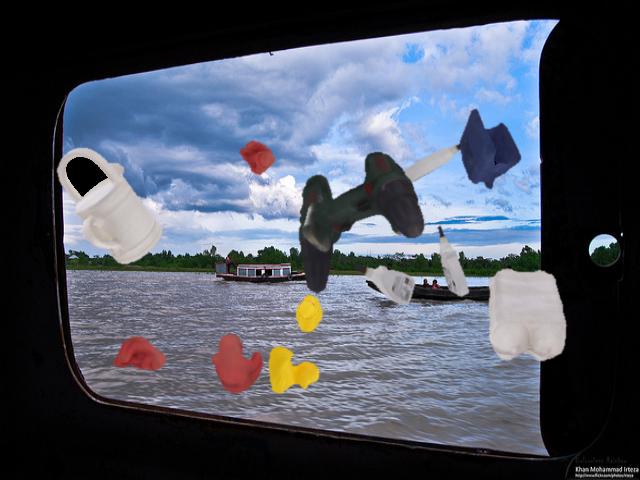}
    \caption{Examples of our synthetic training images.}
    \label{fig:augmented_images}
\end{figure}

\paragraph{Training Set}
For successfully training a deep learning model, the choice of the training data is crucial. In the best case, the training and test data come from the same distribution. Then, it should be relatively easy for the model to generalize from the training to the test images. Unfortunately, not all datasets in the BOP challenge have real labeled training images available.
To train the instance segmentation methods we use real training images whenever provided for a dataset, i.e. TUD-L and YCB-V. For all other datasets we generate synthetic training images. Since we a priori do not know the distribution of test images, we apply the same augmentation strategy for all datasets.
The augmentation is done by cropping the objects from either validation images (e.g. HB) or synthetic training images (e.g. IC-BIN) and pasting them randomly on images from the COCO dataset~\cite{lin2014microsoft}. Thereby we vary the objects' rotation, translation and scale. At most 20 objects are pasted into one image. For each dataset we generate 10000 such images. We use 10 percent of them as validation images and 90 percent as training images. Some example images are shown in Figure~\ref{fig:augmented_images}. Comparing these synthetic images to the real test data, it is obvious that the domain gap is large. Therefore, it is important to avoid overfitting on the training data as much as possible.  During training we additionally apply online augmentation to 70 percent of the samples by either flipping them horizontally or applying color variations. This further increases the variance in the training data and increases the generalization capability of the instance segmentation method.

For each dataset, we also evaluated if including the provided PBR images~\cite{denninger2019blenderproc} further close the domain gap. We choose the final training set based on the mAP on the validation set. Based on this metric, the PBR images are additionally used in training for the datasets LM-O, YCB-V, ITODD and T-LESS. We report the mAP values of the final models on both the validation and the BOP test set in Table~\ref{tab:dl_eval}. For most datasets the gap between validation and test set is significant. It can additionally be seen that the augmentation method is not suited equally for each dataset.

\begin{table}[t]
\begin{center}
\caption{mAP values at IoU threshold 0.5 of the final models evaluated on the validation and BOP test set respectively. Datasets marked with * are evaluated class agnostic, since there is no ground truth information for the test set available.}
\label{tab:dl_eval}
\begin{tabular}{c|cc|cc|cc|cc|cc|cc|cc}
     \bf Dataset & \multicolumn{2}{c}{LM-O} & \multicolumn{2}{c}{T-LESS} & \multicolumn{2}{c}{TUD-L} & \multicolumn{2}{c}{IC-BIN} & \multicolumn{2}{c}{ITODD*} & \multicolumn{2}{c}{HB*} & \multicolumn{2}{c}{YCB-V} \\ \hline
     \bf split & val & test & val & test & val & test & val & test & val & test & val & test & val & test \\
     \hline
     \bf mAP@IoU0.5 & 87.7 & 66.7 & 69.0 & 72.6 & 100.0 & 94.6 & 84.6 & 34.2 & 66.1 & 42.1 & 72.9 & 70.8 & 73.4 & 82.9 
\end{tabular}
\end{center}
\end{table}

\begin{figure}[ht]
    \centering
    \includegraphics[width=0.24\linewidth]{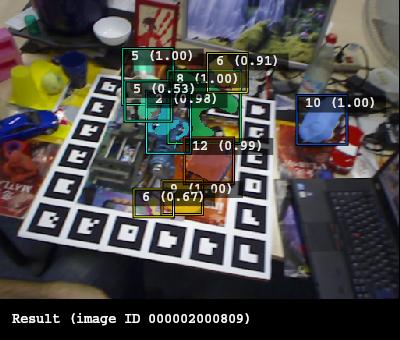}
    \includegraphics[width=0.24\linewidth]{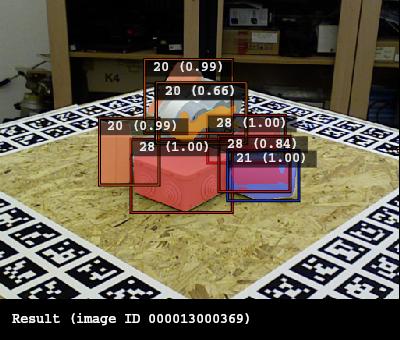}
    \includegraphics[width=0.24\linewidth]{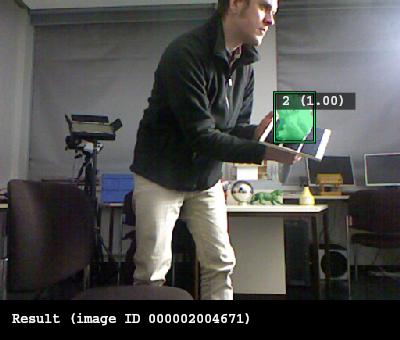}
    \includegraphics[width=0.24\linewidth]{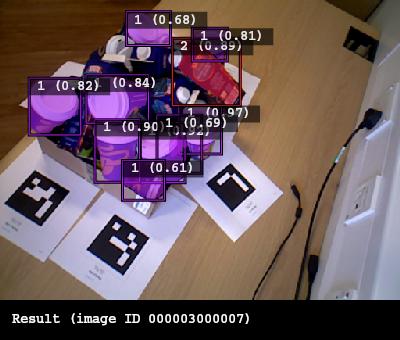} \\
    \includegraphics[width=0.24\linewidth]{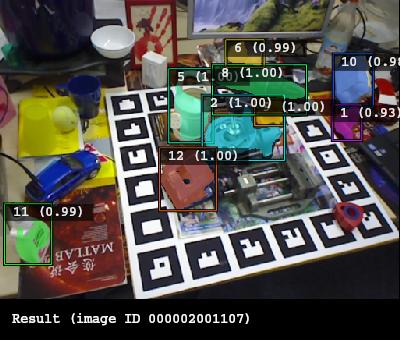}
    \includegraphics[width=0.24\linewidth]{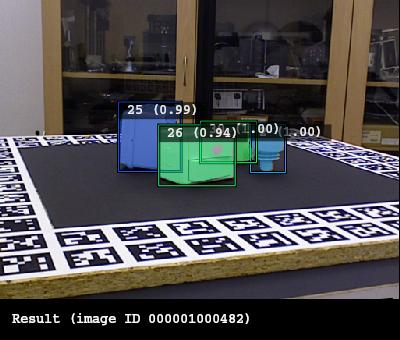}
    \includegraphics[width=0.24\linewidth]{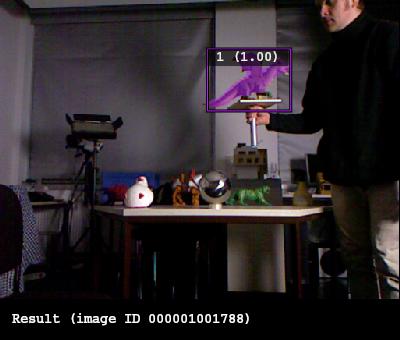}
    \includegraphics[width=0.24\linewidth]{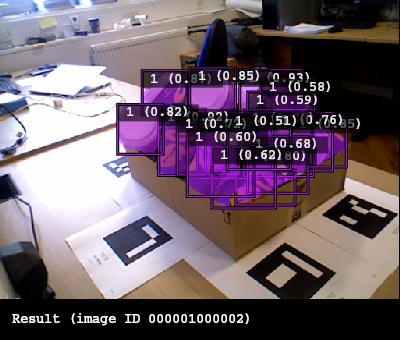} \\
    \includegraphics[width=0.24\linewidth]{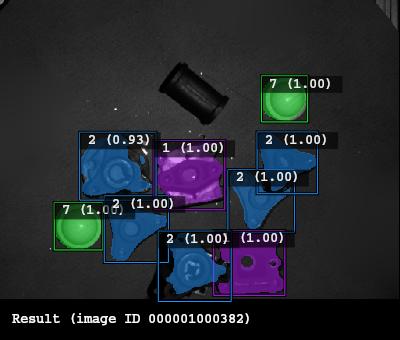}
    \includegraphics[width=0.24\linewidth]{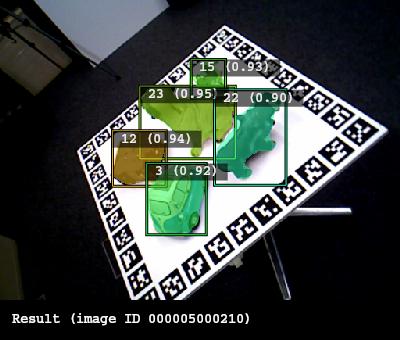}
    \includegraphics[width=0.24\linewidth]{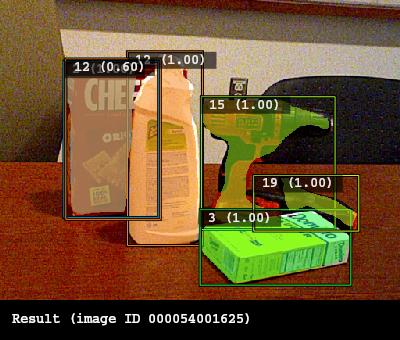} \\
    \includegraphics[width=0.24\linewidth]{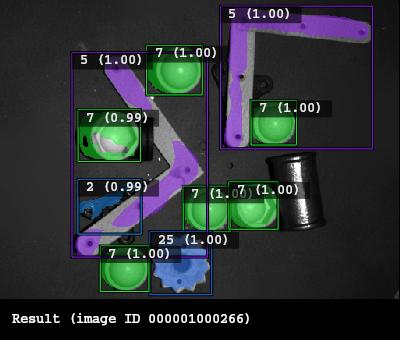}
    \includegraphics[width=0.24\linewidth]{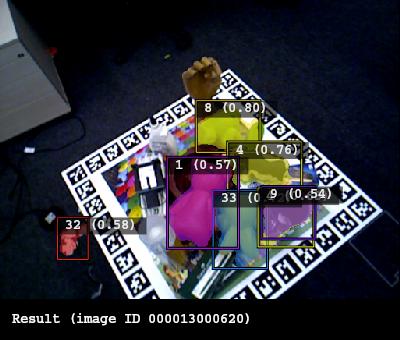}
    \includegraphics[width=0.24\linewidth]{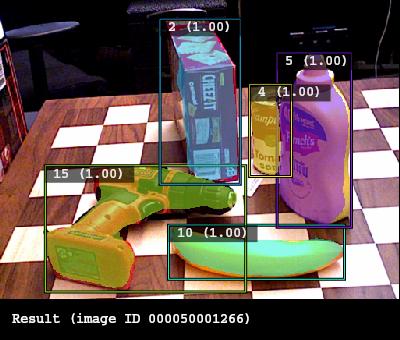}
    \caption{Qualtitative results of instance segmentation on some example images for all datasets.}
    \label{fig:results_instance_segmentation}
\end{figure}

\paragraph{Training Details}
For both Mask-RCNN and RetinaMask we use a ResNet-50~\cite{he2016deep} as backbone with input dimensions 512x384x3 (512x384x1 for ITODD). We also include a Feature Pyramid Network (FPN) \cite{lin2017feature} in the model. The models were pre-trained on the COCO dataset \cite{lin2014microsoft}. As anchor parameters we always use the default setting of three aspect ratios (0.5, 1.0 and 2.0) and three subscales. The minimum and maximum levels of the FPN are automatically determined by the object sizes in the training set. We train the models with Stochastic Gradient Descent (SGD) using an initial learning rate of 0.0001 and a momentum of 0.9 for 20 epochs. For regularization we add a L2-loss on the weights with factor $10^{-5}$. We apply early stopping, i.e. we evaluate at every epoch and choose the model with the best mAP.
Results on some example images are shown in Figure~\ref{fig:results_instance_segmentation}.

\paragraph{Pose Estimation}
Given a set of instance masks and their corresponding object ID, we apply the point-pair voting of \cite{drost2010model} to recover the pose of the object instances, using the implementation in HALCON 20.05 progress \cite{halcon2020a}. Note that this method uses depth only and performs the alignment using only the 3D points and their normal vectors. We experimented with variants of the method that also include edges in the depth or RGB images, but found those to be significantly slower while only marginally improving the results. This is in contrast to the results of the BOP 2019 challenge, where point-pair based methods that also included edges in the voting, refinement, or verification stage significantly exceeded the baseline method. We believe that this is due to the good segmentation from the instance segmentation network.

For datasets where textured objects are available and where the texture is relevant to find the correct pose from a set of symmetric poses (YCB-V), we use a feature-point based approach \cite{lepetit2006keypoint,ozuysal2007fast,halcon2020a} to select the symmetry pose that best matches the instance segmented in the RGB image. After recovering an object pose, the object is rendered at the found location in all symmetric positions and the feature points are extracted. The symmetry with the most matching feature points between rendered object and scene is used as final pose.

\section{Results}

\begin{table}[t]
\begin{center}
\caption{Comparison of the proposed approach with its baseline method, which performs point pair voting only, on the BOP core datasets}
\label{tab:results_vs_baseline}
\begin{tabular}{c|c c c c c c c | c | c}
     {\bf Dataset} & LM-O & T-LESS & TUD-L & IC-BIN & ITODD & HB & YCB-V & avg. & time \\
     \hline
     Voting only~\cite{drost2010model}  &
       $0.527$ & $0.444$ & $0.775$ & $0.388$ & $0.316$ & $0.615$ & $0.344$ & $0.487$ & $7.704 s$ \\
     Ours & 
     $0.631$ & $0.655$ & $0.920$ & $0.430$ & $0.483$ & $0.651$ & $0.701$ & $0.639$ & $0.633 s$ \\
\end{tabular}
\end{center}
\end{table}

\paragraph{Comparison to Baseline}
Compared to the baseline approach~\cite{drost2010model}, which searches the complete scene, the proposed approach is over 12 times faster and has a 15\% higher average recognition rate on the BOP core datasets (see Table~\ref{tab:results_vs_baseline}). While the speedup is due to the reduced search space, the increased recognition rate can be explained by using the RGB images as additional modalities. Since the baseline method uses geometry only, it often finds false positives if clutter is similarly shaped as an object. This is common for objects with large planar sides (such as boxes), which are then found in background planes. The reduction of the search space based on the RGB images effectively avoids this.

\definecolor{background_cell_ours}{rgb}{0.8,1.0,1.0}
\definecolor{background_cell_baseline}{rgb}{0.9,1.0,1.0}

\begin{table}[t]
\begin{center}
\caption{Leaderboard of the BOP 2020 Challenge. The methods are sorted by the average recognition rate over all seven core datasets. The proposed method has the overall rank 2 and is the best method with a runtime faster than 1 second. The baseline method without instance segmentation pre-processing has rank 12. Time is the average runtime per image.}
\begin{tabular}{c l c  c  r}
Rank & Method & Test Modality & $AR_{Core}$ & Time (s) \\
\hline
1 & CosyPose SYNT+REAL-ICP~\cite{labbe2020}  & RGB-D  & 0.698 & 13.74 \\
\rowcolor{background_cell_ours} 2 & Koenig-Hybrid-DL-PointPairs (ours)  & RGB-D  & 0.639 & 0.63 \\
3 & CosyPose SYNT+REAL  & RGB  & 0.637 & 0.44 \\
4 & Pix2Pose-BOP20\_w/ICP-ICCV19~\cite{park2019pix2pose}  & RGB-D  & 0.591 & 4.84 \\
5 & CosyPose PBR & RGB  & 0.570 & 0.47 \\
6 & Vidal-Sensors18~\cite{vidal2018method}  & D  & 0.569 & 3.22 \\
7 & CDPNv2 (RGB-only \& ICP)~\cite{li2019cdpn}  & RGB-D  & 0.568 & 1.46 \\
8 & Drost-CVPR10-Edges  & RGB-D  & 0.550 & 87.56 \\
9 & CDPNv2 (PBR-only \& ICP)  & RGB-D  & 0.534 & 1.49 \\
10 & CDPNv2 (RGB-only)  & RGB  & 0.529 & 0.93 \\
11 & Drost-CVPR10-3D-Edges  & D  & 0.500 & 80.05 \\
\rowcolor{background_cell_baseline} 12 & Drost-CVPR10-3D-Only~\cite{drost2010model} (baseline)  & D  & 0.487 & 7.70 \\
13 & CDPN\_BOP19 (RGB-only)  & RGB  & 0.479 & 0.48 \\
14 & CDPNv2 (PBR-only \& RGB-only)  & RGB  & 0.472 & 0.97
\end{tabular}
\label{tab:results_bop}
\end{center}
\end{table}

\paragraph{BOP Challenge 2020}

The method was submitted to the BOP 2020 challenge where it scored the overall second place (Table~\ref{tab:results_bop}) and was the best performing method with an average runtime of less than 1 second per image. Notably, the introduced method has $7\%$ higher average recognition rate than the winner of the BOP 2019 challenge, while being around 5 times faster.

\section{Conclusion}

We introduced a method that recovers the rigid 3D pose of an object in an RGB-D scene, using a two-stage detector. The first stage is a state of the art, off the shelf instance segmentation network that detects, segments and identifies object instances in the RGB image. The second stage is a vanilla point pair voting scheme that recovers the locally optimal rigid pose. Additionally, we automatically select the best instance segmentation network and training set using the validation error.

The proposed method is fast and robust, and significantly outperforms the baseline method in both runtime and detection performance and is the second best method in the BOP 2020 challenge, and the best with a runtime of less than one second.

%
%

\bibliographystyle{splncs04}
\bibliography{W88P11}
\end{document}